\def\year{2021}\relax
\documentclass[letterpaper]{article} 
\usepackage{aaai21}  
\usepackage{times}  
\usepackage{helvet} 
\usepackage{courier}  
\usepackage[hyphens]{url}  
\usepackage{graphicx} 
\usepackage{amsmath}
\usepackage{amssymb}
\usepackage{booktabs}
\usepackage{program}
\usepackage{natbib}  
\usepackage{caption} 
\usepackage[noabbrev,capitalize]{cleveref}
\usepackage[caption=false,font=footnotesize,labelformat=simple]{subfig}

\usepackage{xcolor}
\usepackage{soul}
\usepackage{acronym}
\acrodef{DL}{deep learning}
\acrodef{ML}{machine learning}
\acrodef{NN}{neural network}
\acrodef{ANN}{artificial neural network}
\acrodef{DNN}{deep neural network}
\acrodef{CNN}{convolutional neural network}
\acrodef{GAN}{generative adversarial network}
\acrodef{PP-GAN}{privacy-preserving \ac{GAN}}
\acrodef{AS}{attack scenario}
\acrodef{LSB}{least significant bit}
\acrodef{DCT}{discrete cosine transform}
\acrodef{FC}{fully connected}

\usepackage[switch]{lineno}
\usepackage{pifont}
\usepackage{listings}

\title{Subverting Privacy-Preserving GANs: Hiding Secrets in Sanitized Images}
\author{
    Kang Liu, Benjamin Tan, Siddharth Garg \\
}
\affiliations{
    \textsuperscript{\rm 1}NYU Tandon School of Engineering\\
    370 Jay Street\\
    Brooklyn, New York 11201\\
    \{kang.liu, benjamin.tan, siddharth.garg\}@nyu.edu
}

\begin{document}

\maketitle

\begin{abstract}
Unprecedented data collection and sharing have exacerbated privacy concerns and led to increasing interest in privacy-preserving tools that remove sensitive attributes from images while maintaining useful information for other tasks. Currently, state-of-the-art approaches use privacy-preserving generative adversarial networks (PP-GANs) for this purpose, for instance, to enable reliable facial expression recognition without leaking users' identity. However, PP-GANs do not offer formal proofs of privacy and instead rely on experimentally measuring information leakage using classification accuracy on the sensitive attributes of deep learning (DL)-based discriminators. In this work, we question the rigor of such checks by subverting existing privacy-preserving GANs for facial expression recognition. We show that it is possible to hide the sensitive identification data in the sanitized output images of such PP-GANs for later extraction, which can even allow for reconstruction of the entire input images, while satisfying privacy checks. We demonstrate our approach via a PP-GAN-based architecture and provide qualitative and quantitative evaluations using two public datasets. Our experimental results raise fundamental questions about the need for more rigorous privacy checks of PP-GANs, and we provide insights into the social impact of these.

\end{abstract}


\section{Introduction\label{sec:01intro}}
The availability of large datasets and high performance computing resources has enabled new \ac{ML} solutions for a range of application domains.
However, as is often the case with transformative technologies, the ubiquity of big data and \ac{ML} raises new data privacy concerns. 
Given the emergence of applications that use personal data, such as facial expression recognition \citep{chen2018vgan} or autonomous driving \citep{xiong_privacy-preserving_2019} one must take care to provide data relevant to the specific application without inadvertently leaking other sensitive information. 
Despite recent legislative efforts to protect personal data privacy---for instance, the General Data Protection Regulation (GDPR) passed by EU---technology must also play a role in safeguarding privacy~\citep{tene_gdpr_2019}. 

\begin{figure}[t]
    \centering
    \includegraphics[width=0.8\columnwidth]{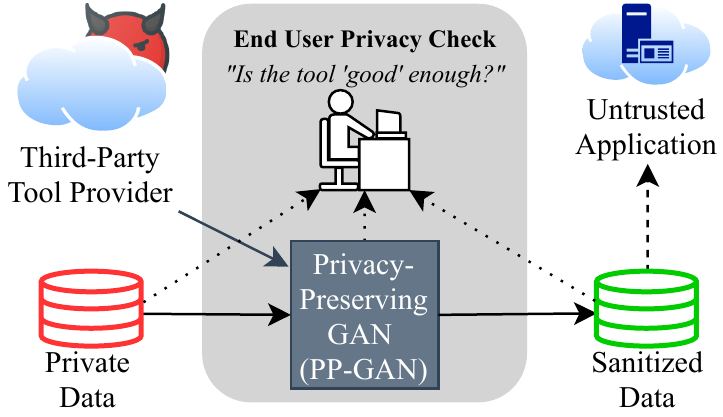}
    \caption{Typical use case for a PP-GAN sourced from a third-party, where a user wants to sanitize their data for use with an (untrusted) application.}
    \label{fig:motivation}
\end{figure}

Consider a scenario where a user wants to use their private data with an untrusted application, as in~\cref{fig:motivation}. 
For privacy, the user needs to remove sensitive---and application irrelevant---attributes from their data while preserving relevant details. 
This can be achieved using a tool typically sourced from a third-party provider. 
To select a tool, the end user performs their own ``privacy check'', evaluating that the tool satisfies their definition of privacy. 

To this end, recent research proposes the use of \acp{DNN}, specifically, \acp{GAN} for sanitizing data of sensitive attributes \citep{wu_privacy-protective-gan_2019,maximov2020ciagan,chen2018vgan}. 
These so-called ``\textit{privacy-preserving GANs}'' (\acused{PP-GAN}\acp{PP-GAN}) can sanitize images of human faces such that only their facial expressions are preserved while other identifying information is replaced~\citep{chen2018vgan}. 
Other examples include: removing location-relevant information from vehicular camera data~\cite{xiong_privacy-preserving_2019}, obfuscating the identify of the person who produced a handwriting sample~\citep{feutry_learning_2020}, and removal of barcodes from images~\citep{raval2017protecting}.  
Given the expertise required to train such models, one expects that users will need to acquire a privacy preservation tool from a third party or outsource \ac{GAN} training, so proper privacy evaluation is paramount. 
In the aforementioned works, researchers note a trade-off between ``utility'' and ``privacy'' objectives---they suggest that \acp{PP-GAN} offer a panacea that achieves both.

The privacy offered by \acp{PP-GAN} is typically measured using empirical metrics of information leakage~\citep{chen2018vgan, xiong_privacy-preserving_2019,feutry_learning_2020}. 
For instance, \citet{chen2018vgan} use the (in)ability of \ac{DL}-based discriminators to identify secret information from sanitized images as the metric for privacy protection. 
However, empirical metrics of this nature are bounded by discriminators' learning capacities and training budgets; we argue that such privacy checks lack rigor. 

This brings us to our paper's motivating question: \textit{are empirical privacy checks sufficient to guarantee protection against private data recovery from data sanitized by a \ac{PP-GAN}?} 
As is common practice in the security community, we answer this question in an adversarial setting. 
We show that \ac{PP-GAN} designs can be subverted to pass privacy checks, while still allowing secret information to be extracted from sanitized images. 
Our results have both foundational and practical implications. 
Foundationally, they establish that stronger privacy checks are needed before \acp{PP-GAN} can be deployed in the real-world. 
From a practical stand-point, our results sound a note of caution against the use of data sanitization tools, and specifically \acp{PP-GAN}, designed by third-parties.
Our contributions include:
\begin{itemize}
    \item We provide the first comprehensive security analysis of privacy-preserving GANs and demonstrate that existing privacy checks are inadequate to detect leakage of sensitive information.
    \item Using a novel steganographic approach, we adversarially modify a state-of-the-art \ac{PP-GAN} to hide a secret (the user ID), from purportedly sanitized face images.  
    \item Our results show that our proposed adversarial PP-GAN can successfully hide sensitive attributes in ``sanitized'' output images that pass privacy checks, with 100\% secret recovery rate. 
\end{itemize}
%
In~\cref{sec:02background}, we provide background on \acp{PP-GAN} and associated empirical privacy checks. 
In~\cref{sec:subvert}, we formulate an attack scenario to ask if empirical privacy checks can be subverted. 
In~\cref{sec:proposed}, we outline our approach for circumventing empirical privacy checks. 
In~\cref{sec:experimental-work}, we present our experimental work. 
In~\cref{sec:discussion}, we discuss our findings in more detail. 
In~\cref{sec:related-work}, we frame our work with reference to prior related work. 
\cref{sec:conclusion} concludes. 

\section{Background\label{sec:02background}}

In this section, we describe the relevant background on our representative \ac{PP-GAN} baseline, how their privacy guarantees are evaluated, and goals of our work. 

\subsection{Representative \ac{PP-GAN} Baseline}
\begin{figure}[t]
    \centering
    \includegraphics[width=\columnwidth]{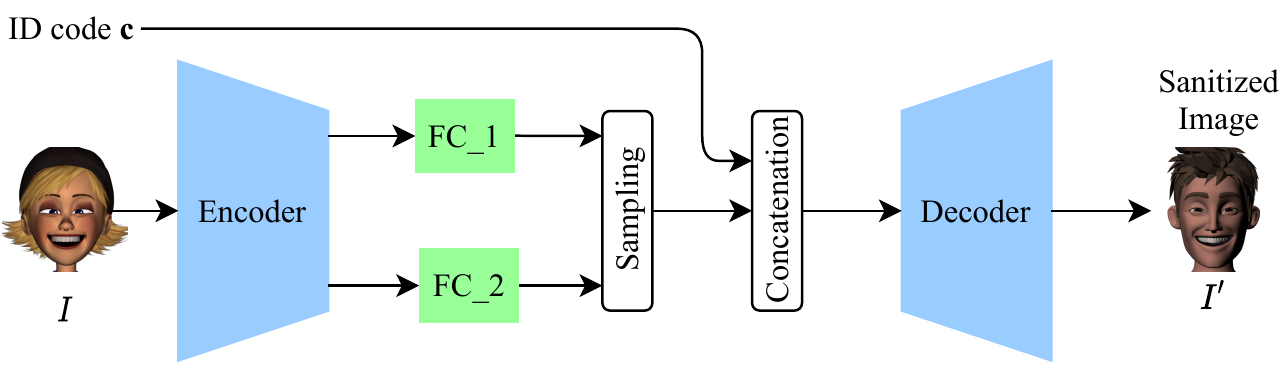}
    \caption{Baseline PP-GAN Architecture}
    \label{fig:orig-arch}
\end{figure}

We adopt the PPRL-VGAN framework proposed by~\citet{chen2018vgan} as our experimental focus\footnote{PPRL-VGAN bears similarities to other related prior works, particularly with respect to privacy checks. We discuss these in more detail in~\cref{sec:related-work}.} and use similar notation. 
PPRL-VGAN produces a \ac{PP-GAN} with a variational autoencoder-generative adversarial network (VAE-GAN) architecture. 
The generator network, $G_{0}$, comprises an encoder, a Gaussian sampling block, and a decoder, as shown in~\cref{fig:orig-arch}. 
The \ac{PP-GAN} is designed to replace the ``user-identity'' in an image while preserving facial expression information. 

Formally, given an input face image $I$ with user ID $y^{id} \in \{0, 1, \cdots, N_{id}-1\}$, expression label $y^{ep} \in \{ 0, 1, \cdots, N_{ep}-1\}$, and a target ID $c$, the generator $G_{0}$ synthesizes a realistic face image $I'$ that belongs to the target ID $c$ while preserving expression $y^{ep}$. 
The user specifies the target identity using a one-hot encoded identity code $c\in\{0,1\}^{N_{id}}$ such that $I' = G_{0}(I,c)$. 
Multiple discriminators, described next, are used to train this \ac{PP-GAN}. 

\subsubsection{Discriminator} %
Unlike conventional \ac{GAN} settings~\citep{goodfellow2014generative}, the training of PPRL-VGAN employs three discriminators, $D^{0}$, $D^{1}$, and $D^{2}$, responsible for real/synthetic face discrimination, ID classification, and facial expression classification, respectively. 
Specifically, $D^{0}$ takes a real or synthetic image as input and outputs the probability of it being real. 
The probability of image $I$ being classified as real is denoted by $D^{0}(I)$. 
Similarly, $D^{1}_{y^{id}}(I)$ and $D^{2}_{y^{ep}}(I)$ denote the probabilities of  $I$ being an image of user $y^{id}$ and having expression $y^{ep}$, respectively.
The discriminators are trained simultaneously 
to maximize the combined loss function $\mathcal{L}_{D}$ which is expressed as:

\begin{equation}\label{eq:D_loss}
    \begin{aligned}
        \mathcal{L}_{D}(D, G_{0}) = & \lambda^{0}_{D}\big(E_{I \sim p_{d}(I)}\log D^{0}(I) + \\ 
        & E_{I \sim p_{d}(I), c \sim p(c)}\log \big(1 - D^{0}\big(G_{0}(I,c)\big)\big)\big) + \\
        & \lambda^{1}_{D} E_{(I, y^{id}) \sim p_{d}(I, y^{id})} \log D^{1}_{y^{id}}(I) + \\
        & \lambda^{2}_{D} E_{(I, y^{ep}) \sim p_{d}(I, y^{ep})} \log D^{2}_{y^{ep}}(I)
    \end{aligned}
\end{equation}

Here, $\lambda^{0}_{D}$, $\lambda^{1}_{D}$ and $\lambda^{2}_{D}$ are scalar constants weighting the different loss components for real/synthetic image discrimination, input image ID recognition, and input image facial expression recognition.

\subsubsection{Generator} 
The generator network has an encoder-decoder architecture. 
The encoder transforms $I$ into two intermediate latents that are fed to a Gaussian sampling block to obtain an ``identity-invariant face image representation'' $f(I)$, i.e., the encoder learns 
a mapping function with random Gaussian sampling, $f(I) \sim q(f(I) \vert I)$. 
The decoder further maps the concatenation of $f(I)$ and $c$ into the synthetic face image $I'$. 

The generator $G_{0}$ aims to generate synthetic face image $I'$ as real as possible to fool discriminator $D^{0}$. 
At the same time, $G_{0}$ learns to synthesize $I'$s that are classified by $D^{1}$ with the target ID specified by $c$. 
In addition, the privacy-protected image $I'$ should maintain the expression $y^{ep}$ from $I$ (as classified by $D^{2}$). 
The combined loss function $\mathcal{L}_{G}(D, G_{0})$ to minimize is: 

\begin{equation}\label{eq:G_loss}
    \begin{aligned}
         & \mathcal{L}_{G}(D, G_{0})= \lambda^{0}_{G}E_{I \sim p_{d}(I), c \sim p(c)}\log D^{0}\big(1-G_{0}(I,c)\big) + \\ 
        & \lambda^{1}_{G} E_{I \sim p_{d}(I), c \sim p(c)} \log D^{1}_{c}\big(1-G_{0}(I,c)\big) + \\
        & \lambda^{2}_{G} E_{(I, y^{ep}) \sim p_{d}(I, y^{ep}), c \sim p(c)} \log D^{2}_{y^{ep}}\big(1-G_{0}(I,c)\big) \\
        & + \lambda^{3}_{G}KL\big(q(f(I) \vert I) \Vert p(f(I))\big)
    \end{aligned}
\end{equation}

The fourth loss term is a \textit{KL} divergence loss used in VAE training that measures the distance between a prior distribution on the latent space $p(f(I)) \sim \mathcal{N}(0, 1)$ and the conditional distribution $q(f(I) \vert I)$. 
Here, $\lambda^{0}_{G}$, $\lambda^{1}_{G}$, $\lambda^{2}_{G}$, and $\lambda^{3}_{G}$ weight the loss components between real/synthetic image discrimination, image ID classification, and expression classification respectively by $D^{0}$, $D^{1}$, and $D^{2}$, and the last \textit{KL} divergence loss.

\subsection{Empirical Privacy Checks}
In the current \ac{PP-GAN} literature, information leakage is measured using empirical privacy checks that quantify the ability of a separately trained \ac{DNN} discriminator to 
pick up trace artifacts post-sanitization that correlate with the sensitive attributes. 
In this work, we focus on two measures based on \acp{AS} proposed by \citet{chen2018vgan}; we refer to these as the ``weak'' and ``strong'' privacy checks. 
For the subsequent discussion, we will assume that the \ac{PP-GAN} is trained using a training dataset of face images and corresponding user IDs and expressions, ${I_{train}, y^{id}_{train}, y^{ep}_{train}}$, and a test dataset ${I_{test}, y^{id}_{test}, y^{ep}_{test}}$ is used to perform the privacy checks.

\subsubsection{Weak privacy check} 
This check corresponds to \ac{AS}~I of Chen et al.'s work, and examines if a discriminator trained on images from the training dataset and their corresponding IDs (i.e., $\{I_{train}, y^{id}_{train}\}$) can recover the original IDs from images $I'_{test} = G_{0}(I_{test},c)$ with $c$ picked at random from $\{0,1\}^{N_{id}}$. 
In other words, if the ID returned by the discriminator given $I'_{test}$ is $y^{id}_{test}$ then the privacy check succeeds. 
We refer to the output test sanitized image as $I'_{test,c}$. 
This check is ``weaker'' than the next check because its discriminator is trained on the distribution of input images and not the distribution of the \ac{PP-GAN}'s sanitized outputs.

\subsubsection{Strong privacy check} 
This check corresponds to \ac{AS}~II of Chen et al.'s work, where the user emulates a stronger adversary and trains a discriminator on sanitized data with the underlying ground-truth identities. 
To address the shortcomings of the weak check, the strong privacy check measures the classification accuracy of a discriminator trained on a dataset obtained by passing training images through $G_{0}$. 
That is, $\{G_{0}(I_{train},c), y^{id}\}$ is used to train the discriminator. 
In other words, the discriminator is trained on the distribution of the \ac{PP-GAN}'s sanitized outputs to recover the original IDs from sanitized test images.

\section{Subverting \acp{PP-GAN}} 
\label{sec:subvert}
\begin{figure}[b]
    \centering
    \includegraphics[width=0.8\columnwidth]{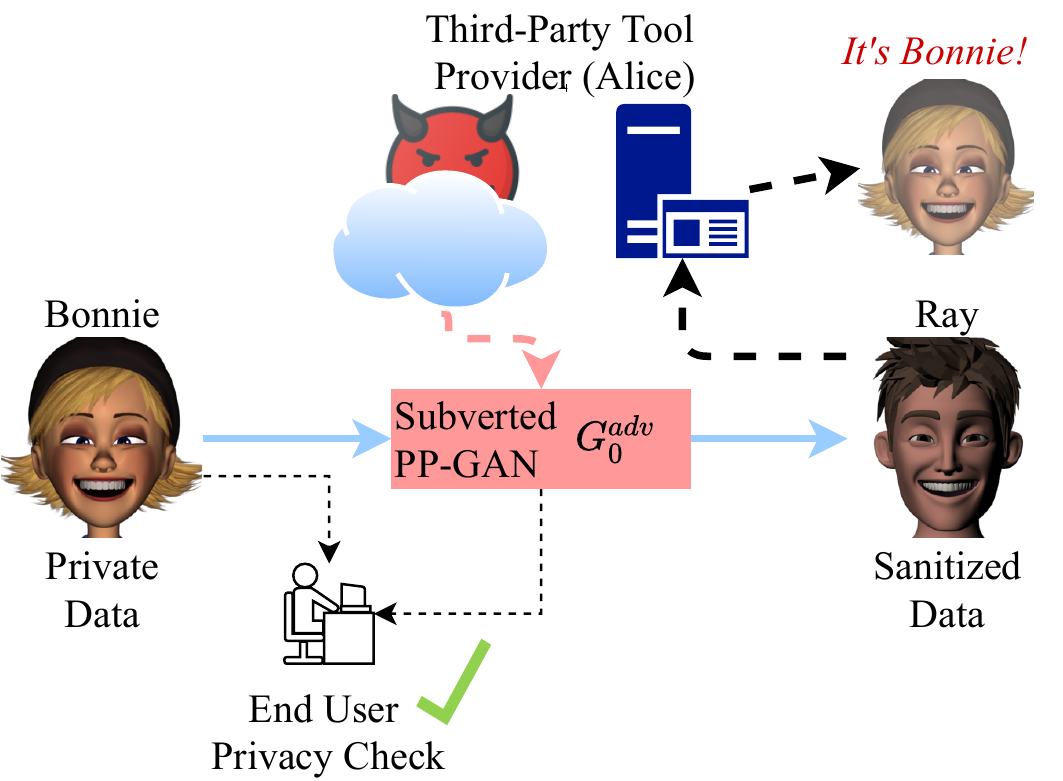}
    \caption{Overview of the Attack Scenario}
    \label{fig:overview}
\end{figure}

We now ask if an adversary can train an adversarial \ac{PP-GAN} $G^{adv}_{0}$ that passes the weak and strong privacy checks, but enables recovery of the sensitive attribute, $y^{id}$, from sanitized outputs $I'$. 
This question reflects the following real-world scenario (\cref{fig:overview}): an adversary, say Alice, trains $G^{adv}_{0}$ and releases it publicly. 
A user, Bonnie, downloads $G^{adv}_{0}$, {\emph verifies} that it passes both weak and strong privacy checks (using validation data), and then uses $G^{adv}_{0}$ to sanitize her private test images and releases them publicly. 
Can Alice (or a collaborator) recover secrets from the sanitized images?

\subsubsection{Goals} Alice seeks to design a \ac{PP-GAN} $G^{adv}_{0}$ with the following goals:

\begin{itemize}
    \item \emph{Utility}: $G^{adv}_{0}$'s outputs, i.e., $I' = G^{adv}_{0}(I, c)$ should have the same expression as that of its input $I$ and $I'$ should be classified as ID $c$;
    \item \emph{Privacy}: the sanitized images should pass both weak and strong privacy checks; and 
    \item \emph{Recovery}: Alice should be able to recover the ID corresponding to image $I$ from the sanitized image $I'$.
\end{itemize}

\subsubsection{Constraints} 
In addition to meeting these goals, Alice wants to ensure that $G^{adv}_{0}$ is still a plausible implementation of a \ac{PP-GAN} to ensure that Bonnie does not identify it as adversarial on inspection (Bonnie is given white-box access to $G^{adv}_{0}$).  
As such, Alice must obey the following:

\begin{itemize}
    \item $G^{adv}_{0}$ still uses the VAE-GAN architecture, i.e., it is a neural network that comprises an encoder, a sampling block, and a decoder; 
    \item $G^{adv}_{0}$ takes in the same inputs as the baseline, which are the input image $I$ and new ID code $c$;
    \item $G^{adv}_{0}$ is allowed to be deeper than the baseline, but the extra layers must be ones that are commonly used in typical \acp{NN}, such as convolutional layers.
\end{itemize}
While the discussion thus far has been in the context of a real-world threat,  an attack that satisfies these goals and constraints has deeper implications on the \acp{PP-GAN} as tools for user privacy preservation. 
Specifically, a successful subversion shows that existing privacy checks are insufficient to fully catch information leakage in \acp{PP-GAN}. 
If one can design an adversarial \ac{PP-GAN} that easily circumvents these privacy checks, it raises questions as to whether information could be leaked via inadvertent design errors as well. 

\section{Proposed Approach\label{sec:proposed}}
We now describe our proposed construction of an adversarial PP-GAN that circumvents both weak and strong privacy checks. 
We begin by discussing a straw man solution and then our implementation.

\subsection{A ``Straw Man'' Solution}
To embed sensitive attributes in sanitized images we turn to steganography, 
a family of techniques that seeks to hide secret information in ordinary files, such as images, without being detected. 
The secret data can then be extracted from the images received by the designated party. 

One possible approach to realize our scheme is to use a conventional steganography tool (e.g., \textit{Steghide}\footnote{\url{http://steghide.sourceforge.net/}}) to embed the user $y^{id}$ in the sanitized face images $I'$. 
In fact, we find that this straw man solution successfully hides the sensitive data. The resultant images pass both privacy checks (results are in \cref{sec:experimental-work}).  
However, recall that we also require our adversarial modifications to be expressed as layers of a \ac{NN}. 
Thus, as we discuss next, we attempt to implement steganographic operations directly. 

\subsection{Adversarial PP-GAN Design}\label{sec:basic}
Implementing steganography in the context of a \ac{PP-GAN} poses several challenges. 
First, the input to an image steganography tool is a ``cover image" \emph{and} the secret; in our setting, the cover image is $I'$ but the secret (user $y^{id}$) is not directly available and must be extracted from $I$. 
For this, we implement a secret extraction stage. 
Second, as noted above, the steganography process must be converted into \ac{NN} layers which poses its own challenges. 
Finally, Alice must be able to extract the secret from the secret-embedded ``sanitized'' image $I''$, which also requires special steps. 
We illustrate the adversarial \ac{PP-GAN} in~\cref{fig:mid-arch} which can subsequently be consolidated to become the final, more innocuous architecture in~\cref{fig:final-arch}. 
After explaining the \ac{NN} building blocks for our approach, we describe the multi-step training process.

\subsubsection{Secret extraction}
The first step is to extract the information pertaining to the secret $y^{id}$ from the input image $I$. 
Since \ac{PP-GAN}'s encoder already has several layers to extract relevant features of $I$, we can extract $y^{id}$ by 
adding an additional \ac{FC} layer FC\_0  in parallel with FC\_1 and FC\_2 (shown as \ding{172} in \cref{fig:mid-arch}). 
The output of FC\_0 is a secret that allows recovery of $y^{id}$. 
In this work, we consider two schemes: \textbf{Scheme 1}---a direct encoding of $y^{id}$ as a one-hot encoded vector and \textbf{Scheme 2}---a vector representation from which the original image $I$ can be reconstructed (and from which $y^{id}$ can be deduced).

\subsubsection{Secret embedding}
\begin{figure}[t]
    \centering
    \includegraphics[width=\columnwidth]{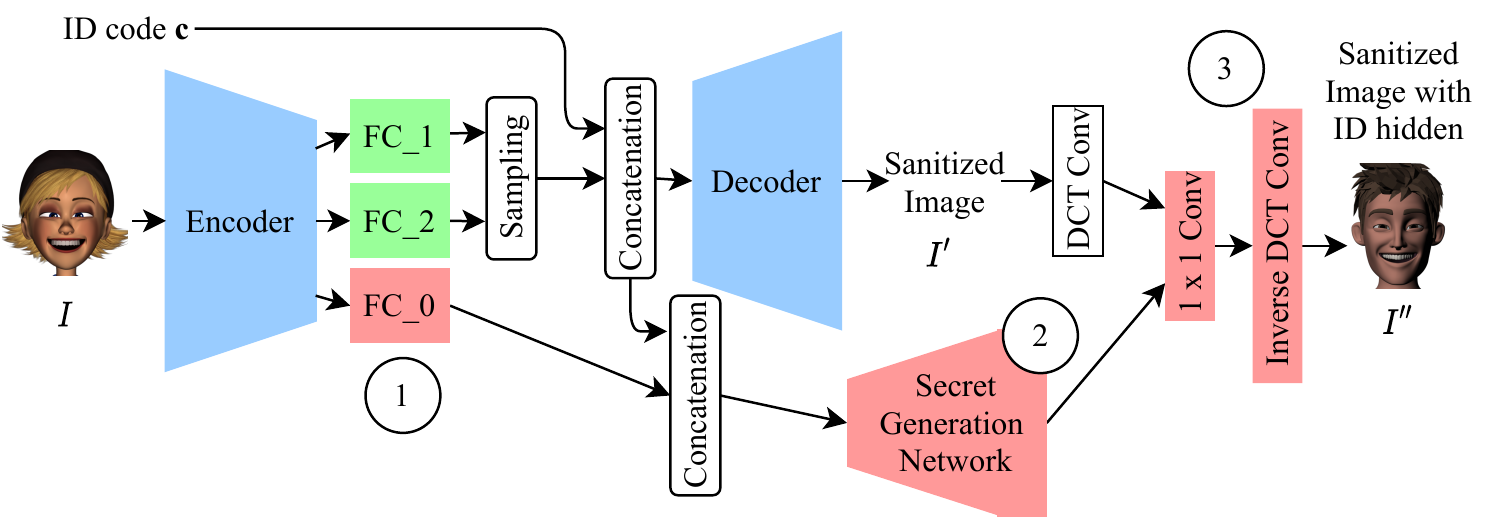}
    \caption{Initial Adversarial \ac{PP-GAN} Architecture}
    \label{fig:mid-arch}
\end{figure}

Our secret embedding method is inspired by steganographic techniques that hide the bits of a secret in the cover image's frequency domain. 
This is achieved by manipulating the \acp{LSB} of the \ac{DCT} coefficients and reduces the visible impact of manipulating an image. 
In our approach, we embed the secret at random locations (\ac{DCT} coefficients) that are \emph{input-dependent}. 
To do so, 
we use $y^{ep}$ and target ID $c$ as 
randomness seeds for selecting the \ac{DCT} coefficients for \ac{LSB} replacement. 
Specifically, $y^{ep}$ is used to select which \ac{DCT} coefficients in which to embed the secret, while $c$ permutes the secret before it is embedded. 
Note that both selection and permutation can be expressed as linear operators and are therefore implementable as \ac{NN} layers (represented as \ding{173} in \cref{fig:mid-arch}). 
As $y^{ep}$ is encoded in $z$, the \ac{PP-GAN} encoder's output latent, and $c$ is explicitly provided, we can simply use bits of their concatenated latent as randomness seeds.

To reduce the impact of \ac{LSB} replacement of \ac{DCT} coefficients in the final image, we embed secrets only in the \acp{LSB} of middle-frequency position \cite{sheisi2012steganography} and select frequency positions that tend to have larger absolute values. 
To determine these locations, we find the average of the absolute values of all middle-frequency \ac{DCT} coefficients in a dataset of (honestly) sanitized images and choose positions with the largest mean. 
Once the secrets are stealthily embedded, the adversarial \ac{PP-GAN} performs inverse \ac{DCT} conversion (\ding{174} in \cref{fig:mid-arch}) to obtain the final ``sanitized'' image, $I''$. 

\subsubsection{Secret Recovery}
The secret recovery stage seeks to extract $y^{id}$ from $I''$. 
Note that secret recovery is not part of the adversarial \ac{PP-GAN} but is a separate process performed by Alice. 
To extract the secret ID information of the user's input image $I$, we first 
classify image $I''$ into expression $y''_{ep}$ and ID $y''_{id}$. 
Assuming that that $y''_{ep} \approx y_{ep}$ and $y''_{id} \approx c$, we can now recover the locations where the secret is embedded and its permutation.
Finally, we perform a \ac{DCT} transformation on $I''$ and extract the secret from the \acp{LSB} of the selected DCT coefficients. 
In scheme 2, where the secret is the vector representation of the original image, Alice can train and use image recovery decoder \ac{NN} to reconstruct $I$.

\subsection{NN-based Implementation}
We now discuss two practical issues with respect to our adversarial \ac{PP-GAN} implementation: ensuring that all functionality is implemented using \ac{NN} layers, and training the final architecture.

\subsubsection{\ac{NN} Layers}
As previously mentioned, the secret extraction leverages existing layers of the \ac{PP-GAN} (encoder) and an additional \ac{FC} layer. 
The secret embedding stage can also be implemented as a multi-layer ``secret generation'' \ac{NN} (\ding{173} in \cref{fig:mid-arch}). 
We use the same architecture as the benign \ac{PP-GAN}'s decoder with an added \ac{DCT} layer at the end; it is trained to permute and embed the secret. 
Finally, \ac{DCT}, addition of LSBs and inverse-DCT can each be implemented using single \ac{NN} layers. \ac{DCT} transformations are simply a linear computation which can be implemented as a convolution, so we manually design these as proposed by \citet{liu2020adversarial}. 
The resulting architecture is indeed a multi-layer \ac{NN} although it has some parallel paths that skip across layers. 
Following training, we can make the adversarial \ac{PP-GAN} appear less suspicious through a merging process using simple transformations and increasing layer dimensions to produce the architecture in \cref{fig:final-arch}. The merging process is described in the Appendix.

\subsubsection{Training} 
Training the adversarial \ac{NN} involves two steps. 
First, we need to train the adversarial \ac{PP-GAN} to extract secret data corresponding to $y^{id}$. 
In scheme 1, we first train the \ac{PP-GAN} with the added FC\_0 layer and introduce an additional loss term to the generator loss function $\mathcal{L}_G$ that measures the $L_2$ distance between the output of the FC\_0 and $y^{id}$ (expressed as a one-hot vector). 
In doing so, the trained network produces both an honestly sanitized image $I'$, as before, but also extracts the secret $y'_{id} \approx y^{id}$. 
In scheme 2, we do the same, but the additional loss term is based instead on the distance between $I$ and the reconstructed image recovered using the output of FC\_0 by an image recover decoder (see below). 
In both schemes, we then freeze the encoder/decoder and FC\_0 weights and focus on the secret generation network to embed the secret data at locations and permutations specified by $y^{ep}$ and $c$. 
We train the secret generation network to minimize the distance of the outputted secret matrix and the manually computed secret matrix using the output of FC\_0, $y^{ep}$ and $c$. 
The discriminators are the same as for the baseline PP-GAN.

In scheme 2, where the secret is a vector representation of $I$, the adversary also trains an image recovery decoder using the same architecture as the decoder in a benign \ac{PP-GAN}. 
The recovery decoder takes as input the output of FC\_0 (which will be the recovered secret). 

\begin{figure}[t]
    \centering
    \includegraphics[width=\columnwidth]{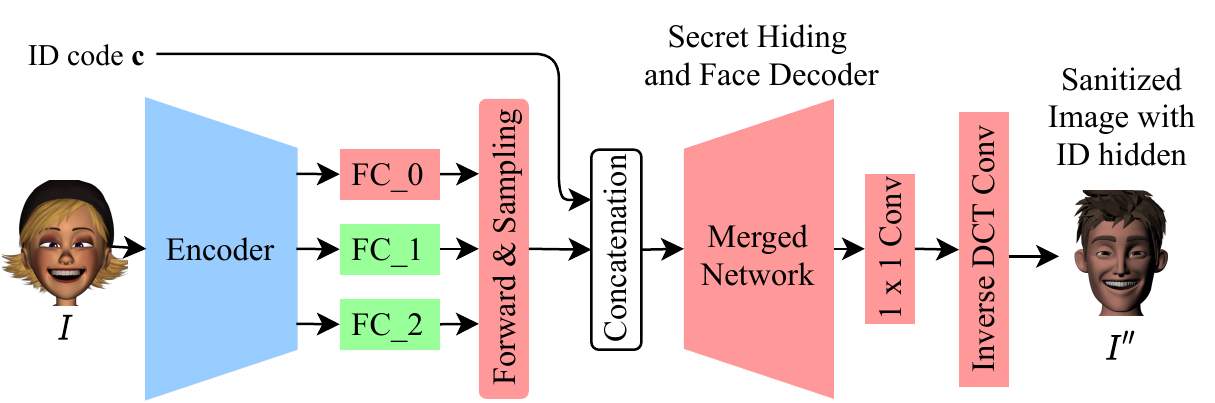} 
    \caption{Proposed Adversarial \ac{PP-GAN} Architecture}
    \label{fig:final-arch}
\end{figure}

\section{Experimental Work \label{sec:experimental-work}}
Details of our experimental setup are in the Appendix. 
\subsection{Datasets}
We validate our proposed approach on two facial expression datasets, FERG~\cite{aneja2016modeling} and MUG~\cite{aifanti2010mug}. 
We split FERG into 47382 training images and 8384 test images with six subjects and seven different expressions. 
For MUG, we select the eight subjects with the most images available. 
Since MUG images are extracted from videos, the initial and final 20 frames in a clip often have neutral expressions so we ignore those frames, resulting in 8795 training images and 1609 test images with seven different expressions. 

\subsection{Evaluation Metrics}
We evaluate the proposed adversarial PP-GAN in terms of three metrics, as described below. 

\subsubsection{Utility Measurement} We measure the utility of the sanitized images via the expression classification accuracy of DL-based discriminators trained on them. Specifically, 
the expression classification accuracy of a discriminator trained on $(I'_{train}, y^{ep})$ and tested on $I'_{test}$ is used as the utility check for the sanitized image dataset. 
Ideally, we would like the utility of data sanitized by the adversarial PP-GAN to be the same as the baseline PP-GAN.

\subsubsection{Privacy Measurement} The privacy of sanitized images is measured using the weak and strong privacy checks described in Section~\ref{sec:02background}. The checks measure the accuracy with which DL-based discriminators can classify the original ID from sanitized face images. Ideally, the adversarial PP-GAN should pass both privacy checks as well as the baseline. 

\subsubsection{Recoverability} Finally, we measure the ability of the adversary to recover the original ID (for scheme 1), the latent representation of the input image (for scheme 2) or the original image itself (for scheme 2) from sanitized face images. The metrics used for each scenario are as follows.

\textit{ID Recovery Accuracy:}  is used for scheme 1 where the adversary seeks to recover the ID of input image $I$ and is defined as the fraction of sanitized test images for which the adversary correctly recovers the secret ID.

\textit{Latent Vector Accuracy:} is used for scheme 2 where the adversary seeks to recover the latent vector corresponding to input image $I$ (from which image $I$ can be reconstructed). Since the latent is a binary vector, we define the latent vector accuracy as the fraction of bits of the recovered latent that agree with the actual latent computed on image $I$.

\textit{Image Reconstruction Error:} is used in scheme 2 to quantify the success of the adversary in reconstructing image $I$ and is defined as the \textit{MSE} distance between reconstructed images and the original input images.

\subsection{Experimental Results}
We begin by discussing our results on the FERG dataset for which the adversary's goal is ID recovery (scheme 1). 
The adversarially trained PP-GAN for FERG has the same utility accuracy as the baseline ($100\%$). Results for the privacy and recoverability metrics are shown in Table~\ref{tab:ferg} for the baseline PP-GAN (Baseline), the adversarial PP-GAN (Adv.), and for the purposes of comparison, a straw man solution in which we use the \emph{Steghide} binary to embed the ID into the baseline PP-GAN's sanitized output.

We observe that the adversarial PP-GAN has the same accuracy for the strong privacy check and only marginally higher accuracy for the weak privacy check compared to the baseline (recall that lower accuracies imply greater privacy). We conclude that the adversarial PP-GAN would therefore pass the privacy checks. At the same time, the adversarial PP-GAN is able to recover the correct ID from sanitized images in all cases. The results from \emph{Steghide} are identical except that it has the same accuracy for the weak privacy check as the baseline. This is because \emph{Steghide} algorithm is fairly sophisticated, but cannot directly be used for our purposes since it is not implemented as an NN.

\begin{figure}[t]
\begin{center}$
\begin{array}{ccc}

I & I' & I''\\
\includegraphics[width=0.6in]{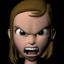}&
\includegraphics[width=0.6in]{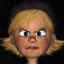}&
\includegraphics[width=0.6in]{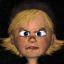}\\

\includegraphics[width=0.6in]{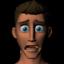}&
\includegraphics[width=0.6in]{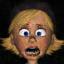}&
\includegraphics[width=0.6in]{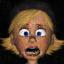}\\

\includegraphics[width=0.6in]{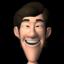}&
\includegraphics[width=0.6in]{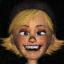}&
\includegraphics[width=0.6in]{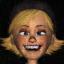}\\

\end{array}$
\end{center}
\caption{Selected Adversarial \ac{PP-GAN} input images $I$, baseline sanitized images $I'$, and outputs $I''$ with hidden secret (FERG dataset)}
\label{Fig:FERG-imgs}
\end{figure}

Figure~\ref{Fig:FERG-imgs} shows examples of images sanitized by the baseline and adversarial PP-GANs (centre and right columns, respectively) along with the input images (left column). Note that the sanitized images produced by the baseline and adversarial networks are visually indistinguishable. 

\begin{figure}[t]
\begin{center}$
\begin{array}{cccc}

I & I' & I'' & I_r\\
\includegraphics[width=0.6in]{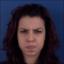}&
\includegraphics[width=0.6in]{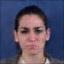}&
\includegraphics[width=0.6in]{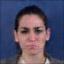}&
\includegraphics[width=0.6in]{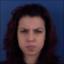}\\

\includegraphics[width=0.6in]{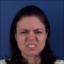}&
\includegraphics[width=0.6in]{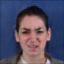}&
\includegraphics[width=0.6in]{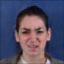}&
\includegraphics[width=0.6in]{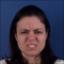}\\

\includegraphics[width=0.6in]{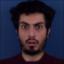}&
\includegraphics[width=0.6in]{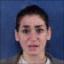}&
\includegraphics[width=0.6in]{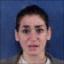}&
\includegraphics[width=0.6in]{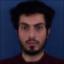}\\

\includegraphics[width=0.6in]{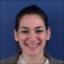}&
\includegraphics[width=0.6in]{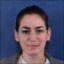}&
\includegraphics[width=0.6in]{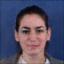}&
\includegraphics[width=0.6in]{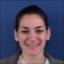}\\

\end{array}$
\end{center}
\caption{Selected Adversarial \ac{PP-GAN} input images $I$, baseline sanitized images $I'$, outputs $I''$ with hidden secret, and reconstructed input images $I_r$ (MUG dataset, 18-bit secrets).}
\label{Fig:MUG-imgs}
\end{figure}


\begin{table}[t]
\centering
\begin{tabular}{@{}lccc@{}}
\toprule
Metric & Baseline & Steghide & Adv. \\ \midrule
Weak Privacy Check          & 0.17     & 0.17  & 0.18      \\
Strong Privacy Check        & 0.30     & 0.30  & 0.30      \\
ID Recovery Acc.  & -        & 1.0   & 1.0      \\ \bottomrule
\end{tabular}
\caption{Privacy and recovery metrics on the FERG dataset for the baseline PP-GAN, using Steghide, and with the proposed adversarial PP-GAN (Adv.).}
\label{tab:ferg}
\end{table}

\begin{table}[t]
\centering
\begin{tabular}{@{}lccc@{}}
\toprule
Metric & Baseline & Steghide & Final \\ \midrule
Weak Privacy Check          & 0.14     & 0.14  & 0.18      \\
Strong Privacy Check        & 0.29     & 0.29  & 0.30      \\
ID Recovery Acc.  & -        & 1.0   & 0.97      \\ \bottomrule
\end{tabular}
\caption{Privacy and recovery metrics on the MUG dataset for the baseline PP-GAN, using Steghide, and with the proposed adversarial PP-GAN (Adv.).}
\label{tab:mug}
\end{table}

\begin{table*}[t]
\centering
\begin{tabular}{@{}lcccccccc@{}}
\toprule
Latent Vector Bit-length    & 18      & 24      & 30      & 36 & 42      & 48      & 54      & 60      \\ \midrule
Strong Check Acc.  & 0.300   & 0.318   & 0.322   &  0.300  & 0.328   & 0.384   & 0.371   & 0.365   \\
Latent Vector Recons. Acc.  & 0.982   & 0.982   & 0.981   &  0.978  & 0.979   & 0.978   & 0.980   & 0.979   \\
Image Recons. Error  & 6.00e-4 & 4.60e-4 & 4.22e-4 &  4.55e-4  & 3.82e-4 & 3.77e-4 & 3.36e-4 & 3.60e-4 \\\bottomrule
\end{tabular}
\caption{MUG dataset, Results after Training Adversarial \ac{PP-GAN} architecture\label{tab:mug-final-results}}
\label{tab:mug2}
\end{table*}

Next we present our results on the MUG dataset for which the adversary's goal is ID recovery (scheme 1) and input image recovery (scheme 2). 
As with FERG dataset, the adversarially trained PP-GANs have the same utility accuracy as the baseline ($100\%$). 

The privacy and recoverability metrics  for ID recovery are shown in Table~\ref{tab:mug}; as with the FERG dataset, we observe that the weak and strong checks on the adversarial PP-GAN have only marginally higher accuracy compared to the baseline, and that adversary is able to recover the correct ID from $97\%$ of sanitized images. Steghide has the same privacy check accuracy as the baseline and $100\%$ recovery rate.

Figure~\ref{Fig:MUG-imgs} shows examples 
of sanitized images for the baseline and adversarial PP-GANs as well as the images recovered by the adversary. The sanitized images from the baseline and adversarial PP-GANs are visually indistinguishable while the recovered images closely resemble the originals. 

Table~\ref{tab:mug2} tabulates the strong privacy check metric, latent vector reconstruction error and image reconstruction error for different latent vector sizes. As larger latent vectors are steganographically embedded in sanitized images, we observe lower reconstruction error at the expense of an increase in the accuracy of the strong privacy check. In all cases, the latent vector is reconstructed with $>97\%$ accuracy. Overall, an 18-bit latent suffices to pass the privacy checks with low reconstruction errors.

\section{Discussion\label{sec:discussion}}
Privacy-preserving GANs have been viewed as somewhat of a panacea to the increasing concerns around surveillance technology. Our results indicate that this view might be too optimistic. Given the concerns that our work raises about the rigor of empirical privacy checks,  there is a need for better evaluations of privacy. 
In this section, we discuss further insights following on from our experimental work. 

\subsection{Secret Hiding}
More secret data bits can be hidden by the adversarial \ac{PP-GAN} trained for the MUG dataset compared to that trained for the FERG dataset. 
This is because real human face data provides more texture variations in pixel-space, which is also reflected in the output sanitized images, and this likely introduces more ``distractions'' for the privacy check discriminators. 
The fact that \ac{DL}-based discriminators are sensitive to such distractions gives us further skepticism into their use as privacy checks. 

Embedding more bits in scheme 2 should help to reduce the image reconstruction error but this is counterbalanced by increasing the difficulty of learning to perform secret embedding. 
If the secret embedding stage is imperfect, there is higher latent vector recovery error and this results in better subversion of the privacy check, as in the case where we hide a 60-bit latent (\cref{tab:mug-final-results}). 

\subsection{\ac{DNN}-based Steganography}
Given our goal of having the adversarial \ac{PP-GAN} implemented entirely as a \ac{NN}, one possible solution is to attach a \ac{DNN}-based steganography tool \citep{zhang2019steganogan, hayes2017generating, zhu2018hidden} to the benign \ac{PP-GAN}. 
The hiding network takes as inputs a cover image and a secret and outputs an image with the secret message hidden. 
The goal of these \ac{DNN}-based steganography approaches is to produce secret-embedded images that are indistinguishable from cover images with respect to the probability that they contain a secret (as measured by empirically a \ac{DL}-based discriminator). 
An accompanying reveal network extracts the secret message from the a secret-embedded image. 
However, this approach is insufficient for our adversarial setting as the resultant network will not pass the strong privacy check. 
As long as there is a DL-based reveal network for secret extraction, we surmise that it is possible to train a DL-based discriminator to classify sanitized images into classes of sensitive attributes.

\subsection{Threats to Validity}
The privacy check measures privacy leakage via classification accuracy of a \ac{DL}-based discriminator on sensitive attributes. Such networks' accuracy is affected by various factors, including the size of the training dataset, network architectures, optimization techniques, weights initialization, and training epochs, \textit{etc}. Thus, the privacy check measurement reported in this paper is only representative of our experimental settings. We would expect different accuracy obtained for a larger dataset, different network architectures, or even more training epochs, but this again points to the unreliability of empirical privacy checks. 

\section{Related Work\label{sec:related-work}}
Conventional privacy-preserving techniques anonymize the sensitive attributes of structured, low-dimensional and static datasets, such as \textit{k-anonymity}~\cite{sweeney2002k} and \textit{l-diversity}~\cite{machanavajjhala2007diversity}. 
\textit{Differential privacy}~\cite{dwork2008differential} was proposed as a more formal privacy guarantee and can be applied to continuous and high-dimensional attributes. 
However, these approaches provide guarantees only when the relationship between sensitive attributes and data samples can be precisely characterized. 

For applications with high-dimensional data, non-sensitive and sensitive attributes intertwine in distributions without a relation model that can be precisely extracted. 
Hence, an empirical, task-dependent privacy checks are used to provide a holistic measure of privacy.
Recent work leverages adversarial networks to sanitize input images and adopts similar \ac{DL}-based discriminators for privacy examination \citep{EdwardsS15, raval2017protecting, pittaluga2019learning, chen2018vgan, wu2018towards, tseng2020compressive,feutry_learning_2020,maximov2020ciagan,xiong_privacy-preserving_2019}. 
These works employs adversarial training to jointly optimize both privacy and utility objectives.  
\citet{EdwardsS15} and~\citet{raval2017protecting} perform simple sanitizing tasks such as removing the QR code from a CIFAR-10 image, or removing the text in a face image, where the sensitive attributes in these cases are artificial and implicit to learn. 
\citet{pittaluga2019learning} learn the privacy preserving encodings via a similar approach but without requiring the sanitized output to be realistic looking. 
Similarly, Wu \textit{et al.} aim to generate degraded versions of the input image to sanitize sensitive attributes. 
The idea of adversarial learning was introduced by~\citet{schmidhuber1992learning}, and motivated \acp{GAN} as proposed by~\citet{goodfellow2014generative}.

\section{Conclusion\label{sec:conclusion}}
Privacy leakage of sanitized images produced by privacy-preserving GANs (PP-GANs) is usually measured empirically using DL-based privacy check discriminators. 
To illustrate the potential shortcomings of such checks, we produced an adversarial PP-GAN that appeared to remove sensitive attributes while maintaining the utility of the sanitized data for a given application. 
While our adversarial PP-GAN passed all privacy checks, it actually hid secret data pertaining to the sensitive attributes, even allowing for reconstruction of the original private image. 
Our experimental results highlighted the insufficiency of existing DL-based privacy checks, and potential risks of using untrusted third-party PP-GAN tools.

\section*{Ethical Impact (Optional)}
Our work critiques empirical privacy check metrics often used in privacy-preserving generative adversarial networks (PP-GANs). 
Given increasing privacy concerns with data sharing, we believe that our work provides timely insights on the implications of such approaches to privacy and will hopefully encourage more work on the rigor of privacy checks. 
Our work describes a technical approach that could allow an adversary to prepare and release an adversarial PP-GAN, although given that PP-GANs are not yet in widespread general use, we expect the negative impact to be minimal.

\bibliography{reference}

\newpage

\begin{table*}[!t]
\centering
\begin{tabular}{@{}clll@{}}
\toprule
Layer & \multicolumn{1}{c}{VAE-GAN Encoder}        & \multicolumn{1}{c}{VAE-GAN/Recovery Decoder}                & \multicolumn{1}{c}{VAE-GAN Discriminator}  \\ \midrule
1     & 5 x 5 x 32 Conv, BN, LeakyReLU  & 2048 FC (4 x 4 x 128), LeakyReLU & 5 x 5 x 32 Conv, BN, LeakyReLU  \\
2     & 5 x 5 x 64 Conv, BN, LeakyReLU  & 5 x 5 x 256 Deconv, BN, LeakyReLU       & 5 x 5 x 64 Conv, BN, LeakyReLU  \\
3     & 5 x 5 x 128 Conv, BN, LeakyReLU & 5 x 5 x 128 Deconv, BN, LeakyReLU       & 5 x 5 x 128 Conv, BN, LeakyReLU \\
4     & 5 x 5 x 256 Conv, BN, LeakyReLU & 5 x 5 x 64 Deconv, BN, LeakyReLU        & 5 x 5 x 256 Conv, BN, LeakyReLU \\
5     & $K$ FC\_0, Sigmoid; 128 FC\_1; 128 FC\_2   & 5 x 5 x 3 Deconv, Tanh                     & 256 FC, LeakyReLU                  \\
6     &                                    &                                            & $D^0$: 1 FC; $D^1$: $N_{id}$ FC; $D^2$: $N_{ep}$ FC   \\ \bottomrule
\end{tabular}
\caption{Adversarial PP-GAN Training Architecture for $K$ bits Secret Extraction}
\label{tab:architecture}
\end{table*}

\section*{Appendix}
\subsection{Network Architectures}
We show in \cref{tab:architecture} the network architectures of different components of adversarial PP-GAN at the secret extraction stage, including encoder, decoder, recovery decoder (used in \textbf{Scheme 2} for input image reconstruction), and discriminators for VAE-GAN training. 

The privacy check discriminator shares the same architecture as the VAE-GAN discriminator except that the last layer is a single FC layer with $N_{id}$ outputs. Similarly, the utility check discriminator is mostly the same as the VAE-GAN discriminator with the last layer being a single FC layer with $N_{ep}$ outputs.

Three parallel DCT convolutional layers following the decoder in \cref{fig:mid-arch} each takes as input one channel of the (honestly) sanitized images $I'$ with the input size being $N \times N \times 1$. Each \ac{DCT} convolutional layer has $n^2$ filters without biases added, and each filter has a size of $n \times n$. The DCT convolutional layer has horizontal and vertical strides of $n$, and the output size for each input channel is $N/n \times N/n \times n^2$. Thus, the concatenated output of all three channels has size $N/n \times N/n \times n^2 \times 3$. This is exactly the \ac{DCT} coefficients of $I'$. In our experiments, $N=64$ and $n=8$. Please refer to \citet{liu2020adversarial} for more details of DCT convolutional layers. 

The secret generation network (\ding{173} in \cref{fig:mid-arch}) uses the same architecture as the decoder followed by the architecture of DCT convolutional layers, and results in a secret matrix of size $N/n \times N/n \times n^2 \times 3$. 

The $1 \times 1$ convolutional layer is applied to the concatenation of the secret matrix and DCT coefficients after reshaping to $N^2/n^2 \times n^2 \times 6$. The output of the $1 \times 1$ convolutional layer has size $N^2/n^2 \times n^2 \times 3$ and is followed by the inverse DCT convolutional layers that have the same architecture as the (forward) DCT convolutional layers.

\subsection{Merging Process for Adversarial PP-GAN NN Implementation}
We transform the two parallel paths of the architecture in \cref{fig:mid-arch} into the proposed adversarial PP-GAN architecture in \cref{fig:final-arch} through a merging process.
Specifically, we merge the secret generation network and the decoding-DCT conversion block into one merged network, made possible because they have identical architectures. 
We merge one \ac{FC} layer using a shared input for both paths, followed by the convolutional layers in both paths. This shared input is the concatenation of FC\_0 output, encoder output $z$, and ID code $c$.
Since fully connected layers and convolutional layers are both linear operations, simple transformation of the weights and biases of layers from both paths will form the merged network that produces outputs with doubled dimensions along the last channel for each layer, i.e., the first half of channels in the merged output is the output from the secret generation network at the corresponding layer, and the second half of channels in the merged output comes from the decoder and DCT conversion block in \cref{fig:mid-arch}.

Here we describe how we transform the weights and biases in both paths into the merged network. 
For the \ac{FC} layer with the shared input $\mbox{concatenate}(L(\text{FC\_0}), z, c)$ in the secret generation network $S$ and decoder-DCT conversion block $T$, the merged output dimension doubles from 2048 as in $S$ and $T$ to 4096 in the merged network $M$, followed by a reshaping layer to generate output with dimension $4 \times 4 \times (128 \times 2)$, which is the input to the following convolutional layers. We refer to the weights of the \ac{FC} layer in $S$, $T$ and $M$ respectively as $w_s$, $w_t$ and $w_m$, and biases as $b_s$, $b_t$ and $b_m$. Here we transform the FC layer weights and biases from $S$ and $T$ to $M$ as the following:
\begin{program}
\BEGIN 
  \FOR j=0 \TO 15 \DO
        $$w_m[:, 256*j:256*j+128] \\
        = w_s[:, 128*j:128*j+128]$$

        $$b_m[256*j:256*j+128] \\
        = b_s[128*j:128*j+128]$$

        $$w_m[:, 256*j+128:256*j+256] = \mbox{concatenate}\\
        ([\mbox{zeros}((K, 128)), w_t[:, 128*j:128*j+128]])$$
        
        $$b_m[256*j+128:256*j+256] \\
        = b_t[128*j:128*j+128]$$
\END
\end{program}
Here, $K$ is the output dimension of FC\_0 (the number of secret bits).

For a specific convolutional layer $i$ in  $S$ or  $T$, its output $L^{S}_{i}$ or $L^{T}_{i}$ has shape $d_{i} \times d_{i} \times c_{i}$, here $c_{i}$ is the number of output channels of layer $i$, and $d_{i}$ is the size of the feature maps. The corresponding layer $i$ in the merged network $M$ has output $L_{i}$ of shape $d_{i} \times d_{i} \times (2 \times c_{i})$, and the output $L_{i}$ has $2 \times c_{i}$ channels. 
To generate such output, the dimensions of the weights at convolutional layer $i$ expands from $f_{i} \times f_{i} \times c_{i} \times c_{i-1}$ as in layer $i$ in $S$ and $T$, to dimensions of $f_{i} \times f_{i} \times (2 \times c_{i}) \times (2 \times c_{i-1})$ in $M$. Here, $f_{i}$ is the size of convolutional filters at layer $i$. Accordingly, the bias dimension expands from $c_{i}$ to $2 \times c_{i}$, from layer $i$ in $S$ and $T$ to $M$. We denote the weights of convolutional layer $i$ in $S$, $T$, and $M$ respectively as $w_{i}^{S}$, $w_{i}^{T}$, and $w_{i}^{M}$, and the biases as $b_{i}^{S}$, $b_{i}^{T}$, and $b_{i}^{M}$. We have the following relationship between convolutional layer weights and biases in networks $S$, $T$, and $M$ to generate the stacked outputs:
\begin{equation}
    \begin{cases}
    w_{i}^{M}[:, :, 0:c_{i}, 0:c_{i-1}] = w_{i}^{S} \\
    w_{i}^{M}[:, :, 0:c_{i}, c_{i-1}:2 \times c_{i-1}] = 0 \\
    w_{i}^{M}[:, :, c_{i}:2 \times c_{i}, 0:c_{i-1}] = 0 \\
    w_{i}^{M}[:, :, c_{i}:2 \times c_{i}, c_{i-1}:2 \times c_{i-1}] = w_{i}^{T} \\
    b_{i}^{M}[0:c_{i}] = b_{i}^{S} \\
    b_{i}^{M}[c_{i}:2 \times c_{i}] = b_{i}^{T} 
    \end{cases}
\end{equation}

The final output of the merged network has six channels, the first three channels are the learned secret matrix, and the last three channels are the \ac{DCT} coefficients of the sanitized image $I'$. 

\subsection{Data Processing}
We resize the original images in the FERG and MUG datasets into dimensions of $64 \times 64$ and normalize the pixel intensities between -1 and 1 as input to the neural networks. The output image pixel values of the VAE decoder and the recovery decoder (used in \textbf{Scheme 2}) are also within -1 and 1. 

In the adversarial PP-GAN, we scale the pixel intensities of the (honestly) sanitized images $I'$ by a factor of $255/2$ and then compute the DCT coefficients, which is followed by a rounding procedure to convert the coefficients to integers. This scaling effect is done by multiplying a constant of $255/2$ to the weights of the DCT convolutional layers. The $1 \times 1$ convolutional layer is essentially a linear combination of the multiple channels of the input data, where the first three channels of the input correspond to the secret matrix (which is also rounded to integer elements) and last last three channels are the scaled DCT coefficients. The $1 \times 1$ convolutional layer multiples 2 to the last three channels and adds that to the first three channels. These DCT convolutional and $1 \times 1$ convolutional layers ensure that the secrets are added to the DCT coefficients of images with pixel values within range -255 and 255 and that these DCT coefficients are rounded to the nearest even number before adding. 
This enesures that the LSBs of DCT coefficients are replaced by the secrets, and allow for efficient secret recovery. One inverse DCT convolutional layer is applied to the LSB-replaced DCT coefficients to generate the final "sanitized" images $I''$ with secrets hidden. The pixel value range of $I''$ is between -255 and 255.   

\subsection{Training Hyper-parameters}
We use the open source implementation from \citet{chen2018vgan} with the same training hyper-parameters to train our adversarial PP-GAN network at the secret extraction stage, which includes training the encoder, decoder, recovery decoder, and discriminators. We train for 300 epochs when using the FERG dataset and 1500 epochs for the MUG dataset (due to smaller training dataset) with RMSprop optimizer and a learning rate of 0.0002. The secret generation network is trained for 1000 epochs with the same optimizer and learning rate and the model with the minimum validation loss is selected. All privacy and utility check discriminators are trained for 1000 epochs with SGD optimizer and a learning rate of 0.05. Models with the highest accuracy are selected for measuring the privacy leakage and utility. Batch size is 256 in all training instances. 

\subsection{Experimental Platform}
We perform NN training/test on Lambda Quad GPU workstation with Intel CPU i9-7920X (12 cores, 2.90 GHz) and Nvidia GeForce GTX 1080 Ti GPUs. We implement our experiments using Keras 2.3.1 and python 2.7 on Ubuntu 18.04.

\subsection{\textit{Steghide} Usage with Command Line}
For our illustrative ``straw man'' solution for hiding secrets in the ``sanitizied'' images, we use a conventional steganography tool, \textit{Steghide} (\url{http://steghide.sourceforge.net}). 
For the experiments in \cref{sec:experimental-work}, we use the following command line options to embed and recover the private information. 

\noindent Secret embedding:
\begin{lstlisting}
  $ steghide embed -ef secret_file -K \
  -N -p "" -cf cover_image_file -sf \
  stego_image_file -v -e none 
\end{lstlisting}

\noindent Secret recovery:	
\begin{lstlisting}[language=bash]
  $ steghide extract -sf \
  stego_image_file -p "" -xf \
  secret_file
\end{lstlisting}

\subsection{Source Code}
Source code will be made available publicly upon publication.
\end{document}